*Article*

# A Novel Proposal in Wind Turbine Blade Failure Detection: An Integrated Approach to Energy Efficiency and Sustainability

Jordan Abarca-Albores [1], Danna Cristina Gutiérrez Cabrera [1], Luis Antonio Salazar-Licea [1],
Dante Ruiz-Robles [1], Jesus Alejandro Franco [1], Alberto-Jesus Perea-Moreno [2,*], David Muñoz-Rodríguez [2]
and Quetzalcoatl Hernandez-Escobedo [1,*]

[1] Escuela Nacional de Estudios Superiores Unidad Juriquilla, UNAM, Querétaro 76230, Mexico;
jordan.abarca@comunidad.unam.mx (J.d.J.A.A.); dagzca@comunidad.unam.mx (D.C.G.C.);
lsalazarlicea@unam.mx (L.A.S.L.); dante_ruiz@unam.mx (D.R.-R.); alejandro.francop@unam.mx (J.A.F.)
[2] Departamento de Física Aplicada, Radiología y Medicina Física, Universidad de Córdoba, Campus Universitario de Rabanales, 14071 Córdoba, Spain; qe2murod@uco.es
* Correspondence: g12pemoa@uco.es (A.-J.P.-M.); qhernandez@unam.mx (Q.H.-E.)

**Abstract:** This paper presents a novel methodology for detecting faults in wind turbine blades using computational learning techniques. The study evaluates two models: the first employs logistic regression, which outperformed neural networks, decision trees, and the naive Bayes method, demonstrating its effectiveness in identifying fault-related patterns. The second model leverages clustering and achieves superior performance in terms of precision and data segmentation. The results indicate that clustering may better capture the underlying data characteristics compared to supervised methods. The proposed methodology offers a new approach to early fault detection in wind turbine blades, highlighting the potential of integrating different computational learning techniques to enhance system reliability. The use of accessible tools like Orange Data Mining underscores the practical application of these advanced solutions within the wind energy sector. Future work will focus on combining these methods to improve detection accuracy further and extend the application of these techniques to other critical components in energy infrastructure.

## 1. Introduction

In recent decades, wind power has emerged as a crucial source of clean and sustainable energy generation [1], playing a key role in reducing greenhouse gas emissions and facilitating the transition to a more sustainable energy matrix. Wind turbines have become essential components of the energy infrastructure, efficiently harnessing wind power to generate electricity in an environmentally friendly manner. Within the operation and maintenance of these turbines, the blades play a crucial role in converting the kinetic energy of the wind into mechanical energy, ultimately transforming it into electricity [2]. However, these massive structures face inherent challenges due to constant exposure to extreme weather forces, dynamic loads, and structural fatigue [3]. Consequently, the blades are susceptible to various types of wear and damage throughout their service life, impacting their performance and, consequently, the efficiency of the entire turbine [4].

The early and accurate detection of wind turbine blade failures has become a crucial priority to ensure the reliability, operational safety, and economic viability of wind farms [5], given that wind turbine blades represent 20% of the total cost of the wind turbine [6]. As the wind industry continues to expand and mature, various innovative techniques and approaches to fault detection have been developed and adopted, opening new opportunities to improve the efficiency and sustainability of wind power generation.

### 1.1. Challenges in Current Detection Methods

A diverse range of methods and tools has been explored, encompassing sensor-based real-time monitoring technologies [7], advanced data analysis approaches [8], and visual inspection techniques [9]. Despite these advancements, there remain significant challenges and gaps:

Sensor Deployment and Data Management: Han and Yang [10] noted the challenges in installing sensors effectively for continuous monitoring. The difficulty lies not only in the installation but also in maintaining these sensors and managing the vast amounts of data they generate, especially in harsh and remote environments where wind turbines are typically located.

Efficient Data Analysis: Zhang et al. [11] proposed a method using mechanical vibrations, reducing the number of samples needed for effective conclusions by 60%. However, the variability in turbine models and operational environments means that the analysis needs to be robust across different conditions, which remains a challenge. Han et al. [12] addressed the scarcity of samples for deep learning, proposing a semi-supervised fault diagnosis model that works efficiently with few samples, yet the model's accuracy under different conditions still requires further validation. In addition, Wang et al. [13] proposed a graph attention autoencoder for blade icing supervision, which utilizes sensor data from supervisory control and data acquisition (SCADA) systems. This method shows promise but requires further testing under various operational conditions to fully validate its effectiveness.

Visual Inspection Innovations: Visual inspection methods, including UAV-based techniques, have advanced significantly. Guo et al. [14] and Bernalte and García [15] utilized UAVs and image analysis for fault detection, highlighting the potential of these methods. However, automating these inspections to reduce human error and ensuring consistent accuracy in various environmental conditions are ongoing issues.

Integration of Acoustic and Vibration Signals: Liu and Zhang [16], Li et al. [17], and Wang et al. [18] proposed combining acoustic emission with vibration signals for detecting cracks and other faults. Although promising, these methods require more research to integrate effectively into comprehensive fault detection systems that can operate reliably in real-world scenarios.

Machine Learning in Fault Detection: Machine learning has shown significant potential, particularly in processing large datasets from SCADA systems. Banala et al. [19] used machine learning for real-time monitoring, and Ogaili et al. [20] proposed an ARIMA model for predictive analysis. Additionally, Rangel et al. [21] integrated convolutional neural networks (CNNs) and frequency–time analysis to detect 62 blade faults with 100% accuracy for unbalanced and bearing faults. Gajbhiye and Warudkar [22] further emphasized the benefits of CNNs in wind turbine blade failure detection, showing that these networks reduce the need for human intervention and lower costs. However, these methods are often data-hungry and computationally intensive, which can be limiting factors in their widespread adoption.

### 1.2. Importance of Energy Efficiency and Sustainability

The relationship between energy efficiency and fault detection in wind turbine blades is well established. Wang et al. [23] demonstrated that predictive maintenance could enhance energy efficiency by reducing downtime and preventing catastrophic failures. Zhang et al. [24] used statistical data from offshore wind farms to show that fault detection could improve energy efficiency, using methods like Fuzzy Fault Tree Analysis to assess risks.

Machine learning, particularly artificial intelligence, has emerged as a crucial tool for fault detection, as exemplified by Mourad et al. [25], who analyzed wind turbines using SCADA data. They implemented a Gaussian mixture model, yielding accurate results in predicting and locating faults. Du et al. [26] proposed, based on data obtained from a SCADA system, a method to differentiate false alarms from real alarms in the detection of faults in wind turbines. To do this, they first use multivariate clustering to obtain appropriate subsets of wind turbines, and then use an autoregressive neural network. Finally, the residuals between the median values of the cluster and the target values are used to calculate the level of anomaly. Choe et al. [27] emphasized the importance of reducing maintenance and operational costs for wind energy system reliability, employing a state monitoring system as an effective tool for wind turbine blade maintenance.

Sustainability considerations are essential in analyzing any renewable energy conversion system. Mourad et al. [25] focused on sustainability in their study, emphasizing the need for fault-free equipment to achieve optimal results in energy transformation. Their evaluation of various phenomena damaging wind turbine blades concluded that maintenance and fault detection could increase blade lifespan, contributing to savings in greenhouse gas emissions. Ogaili et al. [28] proposed vibration analysis methods to enhance sustainability by determining blade failures, including erosion, cracking, and mass imbalance, thereby improving blade performance and reducing maintenance costs while minimizing energy conversion efficiency. Kong et al. [29] determined that monitoring offshore wind turbine blades is crucial for power generation due to substantial structural loads, utilizing non-destructive testing methods, and proposed using UAVs for pinpointing failure locations.

### 1.3. Bibliometric Analysis and Research Clusters

To gain a comprehensive understanding of the current research landscape in wind turbine blade fault detection, a bibliometric analysis was conducted using Scopus and VOSviewer 1.6.20 software. The analysis focused on identifying clusters of related terms and keywords that frequently co-occur in the literature, which reveals the interconnectedness of various research areas.

Figure 1 shows the results of this bibliometric analysis, identifying eight distinct research clusters. These clusters highlight the key areas of focus in the field and their interrelationships:

Advancements in Electrical System Fault Diagnosis (Red): This cluster focuses on topics such as fault diagnosis, vibration analysis, and electric fault currents.

Innovations in Detection and Fault Tolerance in Mechanical Structures (Green): This cluster includes research on cracks, sensors, and fault tolerance in mechanical systems.

Innovations in Wind Turbine Blade Inspection (Blue): Emphasizes the role of inspection, image analysis, and fracture mechanics in improving blade integrity.

Structural Health Monitoring (Yellow): Discusses non-destructive testing methods, including ultrasonic applications for monitoring blade health.

Advancements in Composite Materials Design (Purple): Focuses on the development of composite materials and acoustic emission testing for enhanced blade durability.

Exploring Mechanical Failures in Composite Structures (Cyan): Examines mechanical failures, fatigue testing, and fracture mechanics in composite structures.

Adhesive Joints (Orange): Investigates adhesive technologies and their role in turbine component assembly and maintenance.

Advances in Lightning Protection Systems (Brown): Covers research on lightning protection methods and systems, crucial for turbine safety.

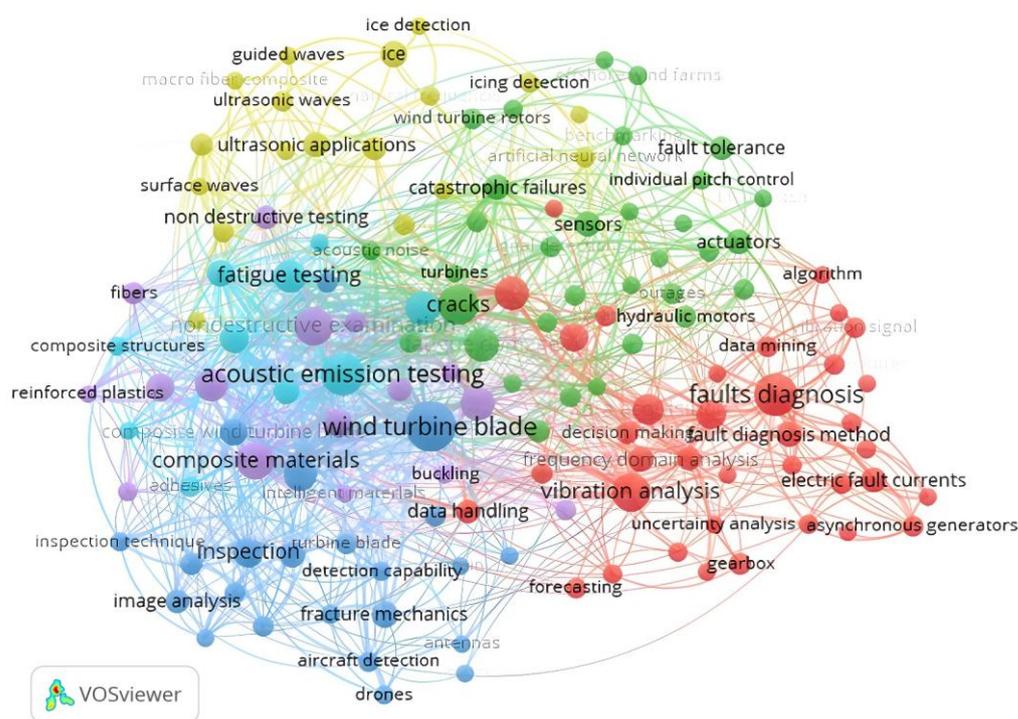

**Figure 1.** Research clusters related to major failure presented in wind turbine blades.

Table 1 presents the primary faults in wind turbine blades, categorized into the eight clusters identified in the bibliometric analysis. This categorization facilitates the observation of relationships between the various research areas, providing a structured overview of the current state of research and highlighting areas where further exploration is needed.

**Table 1.** The main keywords used by the communities detected in the topic Wind Turbine and Failure.

| Cluster | Color | Main Keywords | Topic |
|---------|-------|---------------|-------|
| 1 | Red | Fault diagnosis, Vibration analysis, Frequency domain analysis, Uncertainty analysis, and electric fault currents. | Advancements in Electrical System Fault Diagnosis |
| 2 | Green | Cracks, Sensors, Actuators, Catastrophic failures, and Fault tolerance. | Innovations in Detection and Fault Tolerance in Mechanical Structures |
| 3 | Blue | Inspection, Image analysis, Aircraft detection, Fracture mechanics, Wind turbine blade. | Innovations in Wind Turbine Blade Inspection |
| 4 | Yellow | Ultrasonic applications, Non-destructive testing, Ice detection, Guided waves, and Ultrasonic waves. | Structural health monitoring |
| 5 | Purple | Composite materials, Acoustic emission testing, Adhesives, Intelligent materials, and Reinforced plastics. | Advancements in Composite Materials design |
| 6 | Cian | Failure mechanical, Fatigue testing, composite structures, Fracture mechanics, and Finite element method. | Exploring Mechanical Failures in Composite Structures |
| 7 | Orange | Adhesive, Adhesive thickness, Glass-ceramics, Turbine components, and Composite. | Adhesive Joints |

| 8 | Brown | Lightning, Lightning protection, Lightning strikes, Lightning attachment, and Lightning protection systems. | Advances in Lightning Protection Systems |

This manuscript aims to contribute to the existing knowledge and understanding of wind turbine blade failure detection by addressing gaps identified in the current literature. The integration of sensor data, advanced data analysis techniques, and visual inspection using unmanned aerial vehicles (UAVs), together with a comparative study of different data analysis models, decision trees, naive Bayes, neural networks, logistic regression, and hierarchical clustering, improves the accuracy, efficiency, and sustainability of wind turbine failure detection. The aim of this holistic approach is to optimize the efficiency and durability of wind energy systems, thus facilitating the transition to a cleaner and more sustainable energy source.

## 2. Materials and Methods

Wind power generation plays a crucial role in transitioning to renewable energy sources. However, wind turbines are exposed to adverse environmental conditions that can lead to wear and failure of critical components such as blades. Early detection of these anomalies is essential to ensure the efficient operation and safety of wind turbines.

Wind turbine blades are subject to mechanical stresses and variable weather conditions, which can result in cracks, deformations, or even fractures. Early detection of these failures is essential to avoid costly interruptions in power generation and to ensure the safety of the infrastructure. Conventional inspection techniques often require periodic, manual shutdowns, which entails additional costs and limits the frequency of revisions.

Early detection of blade failures not only contributes to the operational efficiency of wind turbines but also reduces costs associated with unplanned maintenance and extensive repairs. In addition, worker safety is enhanced by minimizing the need for physical inspections in potentially hazardous conditions, such as high altitudes and harsh weather environments.

Currently, techniques and methods for digital image processing are widely used to manipulate, extract, enhance, or represent features from a human perspective. The primary tool used in digital image processing is mathematics, often complemented by various disciplines such as computer science, artificial intelligence, machine learning, computer vision, control, optimization, electrical engineering, and biology. These techniques and methods are implemented in a wide range of applications, including UAVs in aeronautics [30], object observation and exploration in robotics [31], simulation and animation [32], diagnosis [33], and evaluation without the need for invasive methods [34], among others.

In this work, 100 images of wind turbine blades were used, taken from a wind farm located in the Universidad del Istmo area in the Mexican state of Oaxaca. These images include both damaged and undamaged blades.

A digital image is a two-dimensional array or matrix representing the discrete sampling of a continuous signal, where each cell of the matrix is called a pixel and may be constructed of one or multiple channels [35]. Mathematically, such a two-dimensional array can be represented by the Equation (1).

$$f(x,y) \tag{1}$$

where the ordered pair $(x,y)$ represents coordinates in a spatial plane, and $f$ is called the intensity or gray level at any pair of $(x,y)$ coordinates in the image. A two-dimensional image $f(x,y)$ is described by a series of equally spaced samples in the form of an $N * M$ matrix, where each element of the matrix is a discrete quantity according to the pair of $N$ (columns) and $M$ (rows).

Figure 2 shows the position $(x,y)$ in a matrix and its value in three channels that make up the image: RGB (red, green, blue).

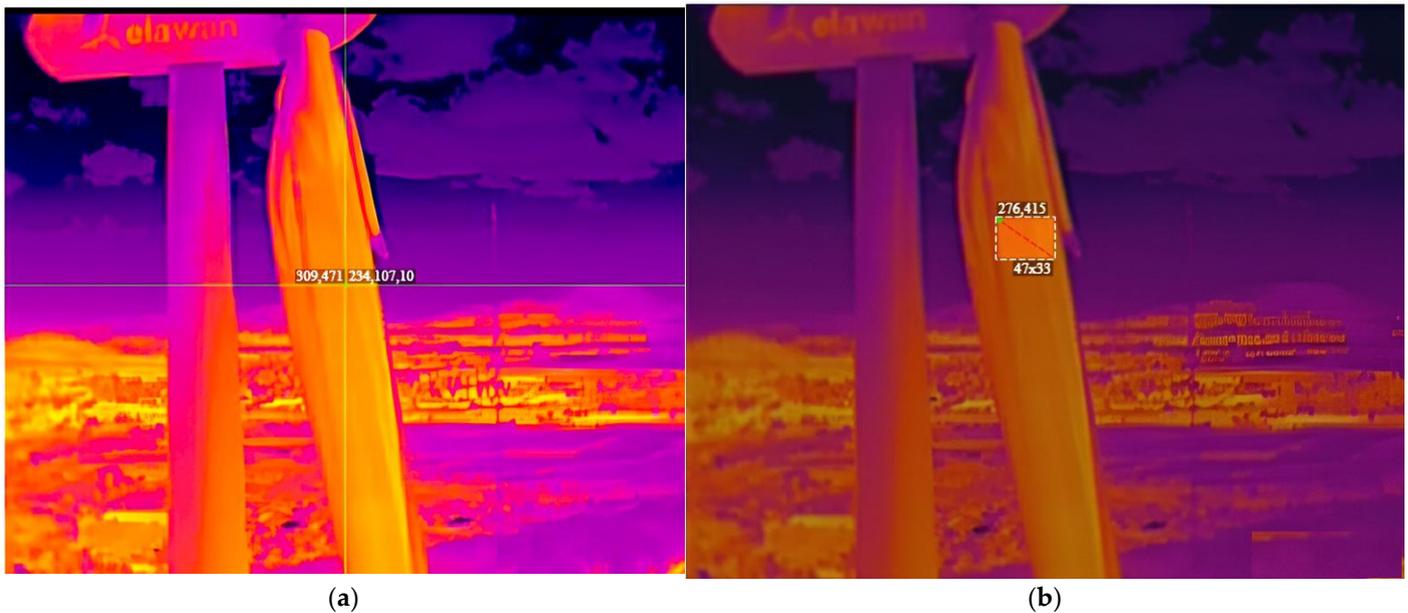

**Figure 2.** Digital image: (**a**) position (*x*,*y*) in a matrix and its value in three channels that make up the image: RGB (red, green, blue); (**b**) area and size of the rectangle.

Thus, the value of each pixel is a discrete number belonging to the domain of positive integers. In a grayscale image, the value of each pixel (resolution) *G* is represented as Equation (2).

$$G = 2^k - 1 \tag{2}$$

Gray levels range between black and white (*G*) in an image of *N* pixels wide by *M* pixels long with *k* bits of resolution. The space required to store the image is calculated as the product of these three variables, i.e., $N \times M \times k$.

## 2.1. Data Mining

The types of supervised machine learning are regression, in which the target variable is continuous, and classification, in which the target variable is categorical. The following is required to build a classification model: characteristics that can be quantified, a labeled target or outcome variable, and a method for measuring similarity.

A linear regression models the relationship between a continuous variable and one or more scaling variables, usually represented as a dependent function equal to the sum of a coefficient plus the scaling factors multiplied by the independent variables. Residuals are defined as the difference between an actual value and a predicted value; a recommended modeling practice for linear regression is as follows: use the cost function to fit the linear regression model; develop multiple models; and compare the results and choose the one that fits the data and whether it is using the model for prediction or interpretation. Three standard measures of error for linear regressions are the sum of squared errors (*SSE*), sum total of squares (*SST*), and coefficient of determination (*R²*).

Unsupervised learning is a machine learning technique in which models analyze unlabeled data to find hidden patterns, relationships, and structures. Unlike supervised learning, where correct answers are provided, unsupervised learning has no human intervention to guide the process. The models discover connections and patterns on their own from the available data [36–39].

In the context of unsupervised learning, algorithms are employed to cluster unlabeled datasets based on their similarities or differences. Some of the more commonly employed approaches include clustering. One of the principal objectives of unsupervised learning is the process of clustering. Clustering techniques are employed to group together data points that are deemed to be similar, based on the similarities or differences that they exhibit. Algorithms such as K-means create clusters by assigning data points to groups (where each group represents a cluster) based on their proximity to a centroid. Clustering is a commonly employed technique in a number of domains, including market segmentation, document grouping, image segmentation, and more. Furthermore, unsupervised learning entails the identification of associations between elements within a given dataset. To illustrate, market basket analysis uncovers the associations between items that are frequently purchased together [40].~~[43].~~ Dimensionality reduction is also a key unsupervised learning technique. A further crucial element is that of dimensionality reduction, which is designed to diminish the number of features or dimensions in the dataset [41]. Techniques such as PCA (principal component analysis) facilitate the simplification of data representation while ensuring the preservation of essential information [42].

*2.2. Image Acquisition*

Image acquisition is a crucial component in developing unsupervised learning projects, as the quality and diversity of the input data directly affect the effectiveness and robustness of the resulting models.

This research focuses on obtaining a diversified and representative dataset, which is essential for training models that can effectively generalize to new situations and contexts. Through a combination of data sources, including in situ captured images and collections from existing databases, a robust dataset has been created to drive this project.

Additionally, the technical and logistical aspects of image acquisition, such as image resolution, lighting conditions, and subject variability, are discussed as fundamental factors to ensure data quality. Image preprocessing techniques, such as normalization and noise reduction, which prepare the data for subsequent analysis, are also addressed.

With the growing processing capacity and access to large volumes of data, unsupervised learning has the potential to uncover complex patterns and relationships in the data that might not be evident otherwise.

The images were acquired using an unmanned aerial vehicle equipped with a ZENMUSE H20 thermal camera that can take one RGB photograph and one thermal photograph; see Figure 3.

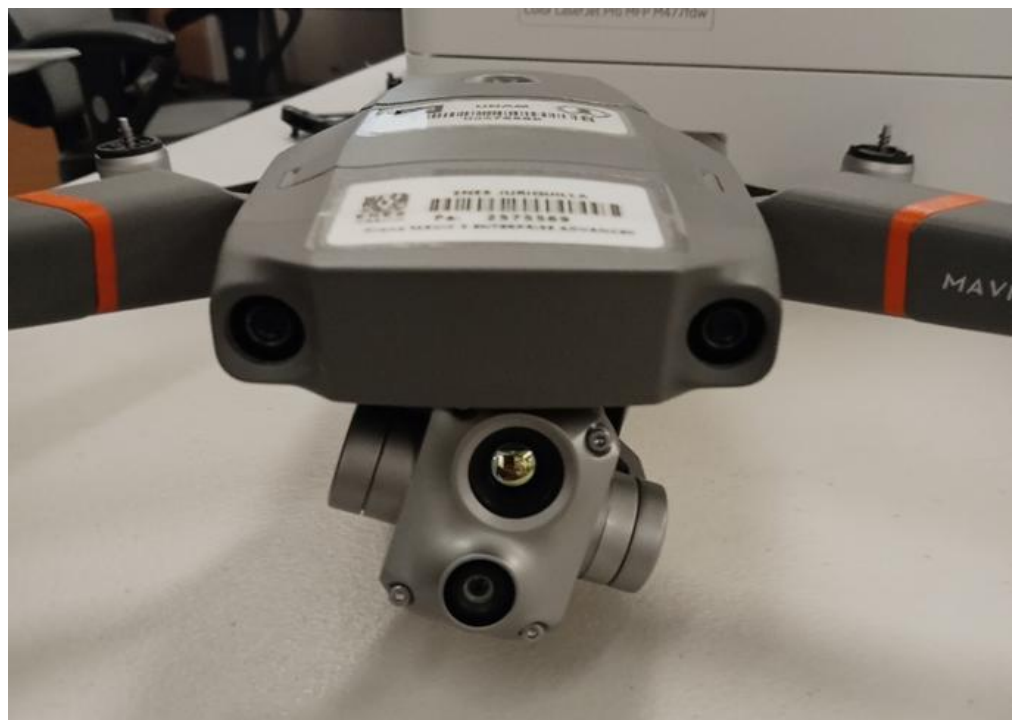

**Figure 3.** ZENMUSE H20 thermal camera.

### 2.3. Orange Data Mining Application and Image Management

Orange is a machine learning and data mining suite for data analysis through Python version 3.12.6 scripting and visual programming [29]. Orange is open-source data mining software version 3.37.0 for visualizing, analyzing, and modeling data.

Several authors, such as Mohapatra and Swarnkar [43], have used Orange Data Mining to compare different artificial intelligence techniques. Ishak et al. [44] used the software in the classification method on the dataset lenses. Tebala and Marino [45] classified sector of economy activity.

Orange provides a graphical user interface that allows users to create data mining workflows by dragging and dropping widgets onto a palette. Widgets can include a simple data table, graph, machine learning model, or custom interface.

The software can perform tasks such as data preprocessing, classification, regression, clustering, network analysis, data visualization, and more. It also supports many data file formats, including CSV, Excel, SQL, and others.

The Orange Data Mining interface has widgets associated with different actions, such as data manipulation, transform, and model, which can be seen in the image analytics. The image embedding widget connects to the Test and Score widget to obtain results of testing classification algorithms. The widget does two things. First, it shows a table with different classifier performance measures, such as classification accuracy and area under the curve. Second, it outputs evaluation results, which can be used by other widgets for analyzing the performance of classifiers, as shown in Figure 4.

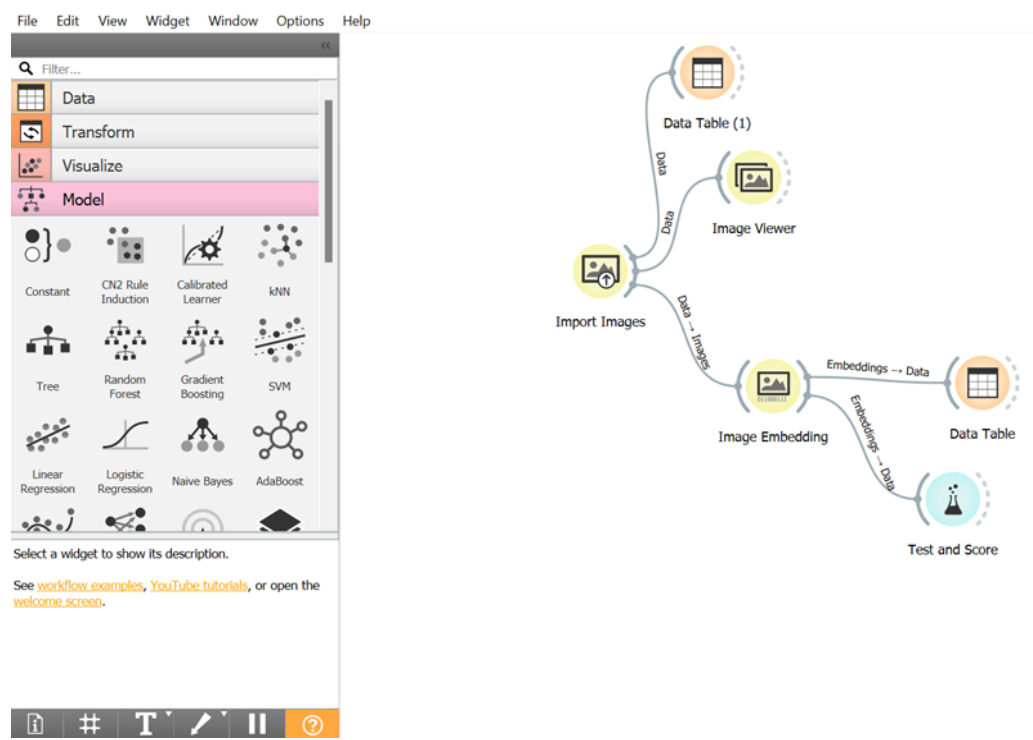

**Figure 4.** Connecting to the Test and Score widget.

It is necessary to use models such as logistic regression, which is a widely used statistical method, for modeling the probability of a binary outcome based on one or more predictor variables. Unlike linear regression, which predicts a continuous outcome, logistic regression is used when the dependent variable is categorical and typically binary (0 or 1, success or failure, respectively) [46]. The core idea of logistic regression is to model the log-odds of the probability of an event occurring as a linear combination of the predictor variables. The log-odds, also known as the logit, is defined as in Equation (3).

$$logit(P) = log\left(\frac{P}{1-P}\right)$$

(3)

where $P$ is the probability of the event occurring. The logistic regression model can be expressed as Equation (4).

$$logit(P) = \beta_0 + \beta_1 X_1 + \beta_2 X_2 + \cdots + \beta_n X_n$$

(4)

Here, $\beta_0$ is the intercept, and $\beta_1$, $\beta_2$, …, $\beta_n$ are the coefficients of the predictor variables $X_1$, $X_2$, …, $X_n$, respectively.

The probability $P$ that the event occurs can be derived from the logit model by applying the inverse of the logit function, which is the logistic function, Equation (5).

$$P = \frac{1}{1 + e^{-(\beta_0 + \beta_1 X_1 + \beta_2 X_2 + \cdots + \beta_n X_n)}}$$

(5)

This transformation ensures that the predicted probabilities lie within the (0, 1) interval. The coefficients $\beta_i$ are typically estimated using maximum likelihood estimation (MLE), which finds the values that maximize the likelihood of the observed data given the model. According to Li et al. [47], tree-based models, such as decision trees, are powerful and interpretable methods for both classification and regression tasks. These models partition the feature space into a set of rectangles (for regression) or regions (for classification) based on the values of input variables, leading to predictions that are easily interpretable and often robust to outliers.

A decision tree model is constructed by recursively splitting the data into subsets based on a feature that maximizes a specific criterion. For a classification problem, the most common criteria are Gini impurity and entropy. The Gini impurity for a node $t$ is given by Equation (6).

$$Gini\ (t) = 1 - \sum_{k=1}^{K} p_k^2$$

(6)

where $p_k$ is the proportion of samples belonging to class $k$ in node $t$, and $K$ is the number of classes. The goal is to choose splits that minimize the weighted sum of Gini impurities of the child nodes.

For regression tasks, the decision tree algorithm typically uses the mean squared error (MSE) as the splitting criterion. The MSE for a node $t$ is given by Equation (7).

$$MSE\ (t) = \frac{1}{N_t} \sum_{i \in t} (y_i - \bar{y}_t)^2$$

(7)

where $y_i$ is the actual value of the target variable for observation $i$, $y_t$ is the mean of the target values in node $t$, and $N_t$ is the number of observations in node $t$. The algorithm selects the split that minimizes the sum of the MSEs of the resulting child nodes.

The splitting process continues until a stopping criterion is met, such as a maximum tree depth or a minimum number of samples per leaf. The resulting tree model can be visualized as a hierarchical structure where each internal node represents a decision based on a feature, and each leaf node represents a predicted outcome.

One of the key strengths of tree-based models is their ability to capture non-linear relationships between the input features and the target variable without requiring feature scaling. Moreover, tree models are naturally capable of handling both numerical and categorical data. However, they tend to be prone to overfitting, which can be mitigated through techniques like pruning or by using ensemble methods such as random forests or gradient boosting; naive Bayes is a simple yet effective probabilistic classifier based on Bayes' theorem, with the strong assumption that features are conditionally independent given the class label [48]. Despite its simplicity, naive Bayes often performs surprisingly well in a variety of classification tasks, particularly in text classification and other domains where this independence assumption holds approximately true. Bayes' theorem forms the foundation of naive Bayes and is expressed as Equation (8).

$$P\,(C|X) = \frac{P\,(X|C) \cdot P(C)}{P(X)} \tag{8}$$

where:

$P\,(C|X)$ is the posterior probability of the class $C$ given the feature vector $X = \{x_1, x_2, \ldots, x_n\}$

$P\,(X|C)$ is the likelihood of observing the feature vector $X$ given the class $C$.

$P(C)$ is the prior probability of the class $C$.

$P(X)$ is the marginal likelihood, which acts as a normalization constant.

Finally, neural networks are a class of machine learning models inspired by the structure and function of biological neurons. These models are particularly powerful in capturing complex, non-linear relationships between inputs and outputs, making them highly suitable for a wide range of predictive tasks [49]. The architecture of a neural network consists of interconnected layers of artificial neurons, typically organized into an input layer, one or more hidden layers, and an output layer.

The output of each neuron in a layer is computed as a weighted sum of its inputs, followed by the application of an activation function. Mathematically, for a given neuron $j$ in layer $l$, the output $a_j^{(l)}$ is given by Equation (9).

$$a_j^{(l)} = f\left(\sum_{i=1}^{n} w_{ij}^{(l)} a_i^{(l-1)} + b_j^{(l)}\right) \tag{9}$$

where $w_{ij}^{(l)}$ represents the weights connecting neuron $i$ in layer $l-1$ to neuron $j$ in layer $l$; $a_i^{(l-1)}$ is the output of neuron $i$ in the previous layer; $b_j^{(l)}$ is the bias term for neuron $j$; and $f(\cdot)$ is the activation function, such as the ReLU (Rectified Linear Unit), sigmoid, or tanh function.

The network's final output is produced by the neurons in the output layer, and the overall network function can be represented as Equation (10).

$$\hat{y} = F(X; \theta) \tag{10}$$

where $\hat{y}$ is the predicted output, $X$ is the input feature vector, and $\theta$ represents the set of all network parameters (weights and biases).

Training a neural network involves finding the optimal parameters $\theta$ that minimize a loss function $L(\hat{y}, y)$, which measures the difference between the predicted output $\hat{y}$ and the true output $y$. This is typically achieved through backpropagation, where the gradients of the loss function with respect to the network parameters are computed using the chain rule, and the parameters are updated via an optimization algorithm such as stochastic gradient descent (SGD); see Equation (11).

$$\theta \leftarrow \theta - \eta \nabla_\theta L(\hat{y}, y) \tag{11}$$

where $\eta$ is the learning rate.

To determine which neural network model fits best for a particular task, one must evaluate several architectures and hyperparameter configurations. Key factors to consider include the number of hidden layers, the number of neurons per layer, the choice of activation functions, and the regularization techniques (e.g., dropout, L2 regularization) to prevent overfitting.

Model selection can be systematically approached by dividing the dataset into training, validation, and test sets. The training set is used to fit the model, the validation set is used to tune hyperparameters and select the best model, and the test set provides an unbiased evaluation of the model's performance. Cross-validation techniques can further enhance the robustness of model selection by averaging performance metrics over multiple folds of the data.

Finally, metrics such as accuracy, precision, recall, F1-score, or mean squared error (depending on the task) are used to compare the performance of different models. The model with the best performance on the validation or test set is considered the best fit for the given problem.

The models employed by Orange Data Mining utilize a neural network to ascertain the optimal model, with the logistic regression widget selected to facilitate this determination. The tree algorithm is a straightforward approach that partitions the data into nodes based on class purity (information gain for categorical and MSE for numeric target variables). The Tree widget, designed in-house, is capable of handling both categorical and numeric datasets. The Neural Network widget utilizes the sklearn Multi-layer Perceptron algorithm, which is capable of learning both non-linear and linear models. Naive Bayes is a fast and simple probabilistic classifier based on Bayes' theorem, assuming feature independence. Subsequently, the confusion matrix provides the number or proportion of instances between the predicted and actual class. The selection of elements from the matrix is then fed into the corresponding instances of the output signal. This allows the user to identify which instances were misclassified and how. Subsequently, an image viewer is incorporated to facilitate observation of the algorithmic identification of images exhibiting defects, as illustrated in Figure 5.

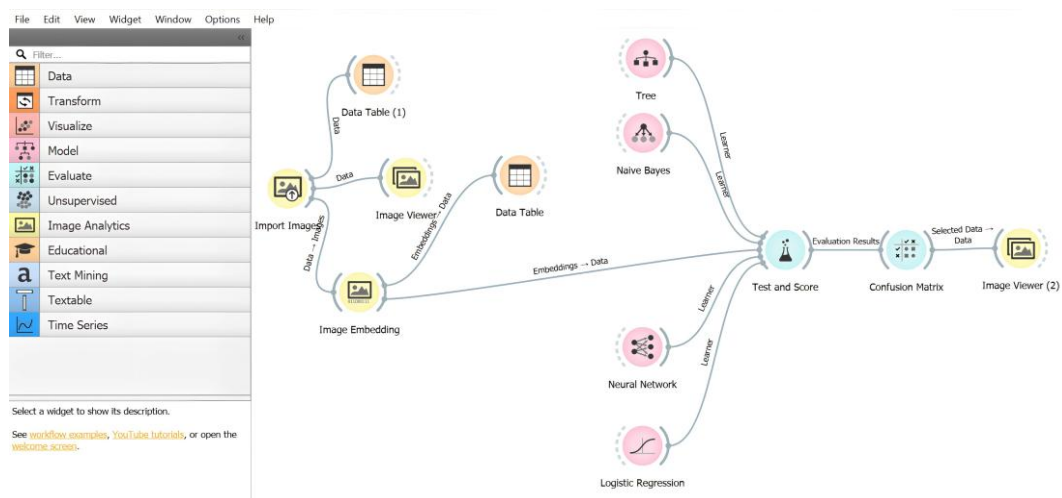

**Figure 5.** Final setup for detecting defects in wind turbine blades.

### 2.4. Clustering

According to Yang et al. [30], deep image clustering networks can categorize unlabeled images, and Orange Data Mining associates the images with their distances. In this case, the Distances widget computes distances between rows or columns in a dataset. The data will be normalized by default to ensure equal treatment of individual features. Normalization is always done column-wise. To visualize the computed clusters, an Image Viewer widget is added. A Hierarchical Clustering widget is used for the hierarchical algorithm. The complete clustering diagram can be seen in Figure 6.

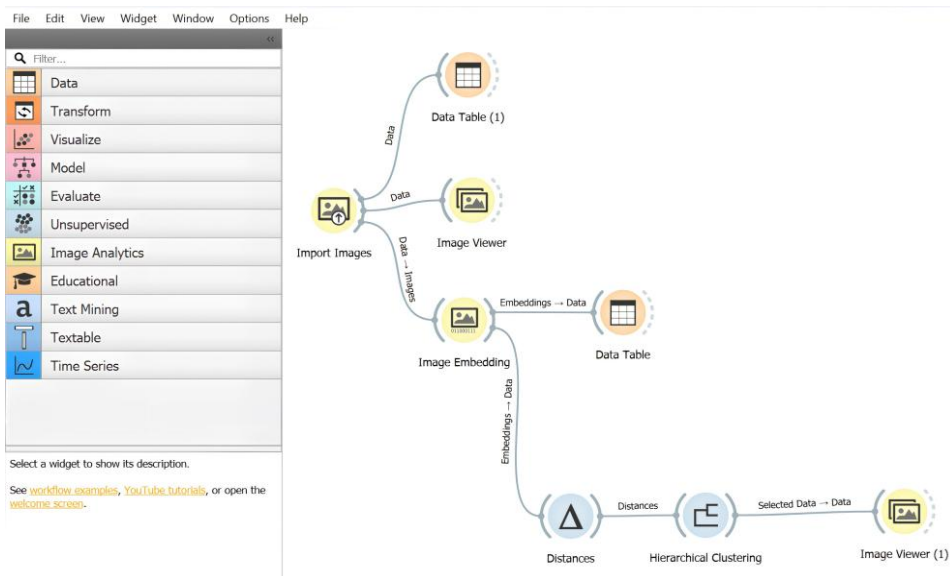

**Figure 6.** Complete clustering diagram.

## 3. Results

Image handling for detecting defects in wind turbine blades is crucial in the wind energy industry. This study demonstrates a viable approach using the Orange Data Mining software, which produced accurate results with image embedding and clustering techniques.

### 3.1. Algorithms Analysis

Four models were evaluated using the "Test Learning Algorithms on Data" functionality. This can be connected to more than one widget to test multiple learners with the same procedures, and it will compute several performance statistics as seen in Table 2.

**Table 2.** Evaluation of the learning models.

| Evaluation Results of Target | | | | | | |
|---|---|---|---|---|---|---|
| **Model** | **AUC** | **CA** | **F1** | **Precision** | **Recall** | **MCC** |
| Tree | 0.783 | 0.708 | 0.719 | 0.742 | 0.708 | 0.554 |
| Naive Bayes | 0.839 | 0.542 | 0.534 | 0.730 | 0.542 | 0.401 |
| Logistic Regression | 0.878 | 0.708 | 0.692 | 0.688 | 0.708 | 0.530 |
| Neural Network | 0.867 | 0.750 | 0.754 | 0.773 | 0.750 | 0.620 |

As shown in Table 2, the learning model that fits the best is logistic regression, which has a value of 0.878 in the area under the ROC (AUC) evaluation, which is defined as the area under the receiver operating curve. Classification accuracy (CA) is the proportion of correctly classified examples. F-1 is a weighted harmonic mean of precision and recall. Precision is the proportion of true positives among instances classified as positive. Recall is the proportion of true positives among all positive instances in the data. The Matthews correlation coefficient (MCC) takes into account true and false positives and negatives and is generally regarded as a balanced measure which can be used even if the classes are of very different sizes.

The results are presented below when comparing the models according to the area under the ROC curve, classification accuracy, F1, precision, recall, specificity, and logistic loss; see Tables 3–9, respectively.

Table 3 shows the probabilities that the score for the model in the row is higher than that of the model in the column; small numbers show that the probability of the difference is negligible.

**Table 3.** Area under ROC curve.

| | Tree | Naive Bayes | Neural Network | Logistic Regression |
|---|---|---|---|---|
| **Tree** | | 0.347 | 0.260 | 0.209 |
| **Naive Bayes** | 0.653 | | 0.250 | 0.223 |
| **Neural Network** | 0.740 | 0.750 | | 0.271 |
| **Logistic Regression** | 0.791 | 0.777 | 0.729 | |

The model comparison shown in Table 3 shows that the logistic regression model once again provides the best fit.

**Table 4.** Classification accuracy.

| | Tree | Naive Bayes | Neural Network | Logistic Regression |
|---|---|---|---|---|
| **Tree** | | 0.893 | 0.439 | 0.526 |
| **Naive Bayes** | 0.107 | | 0.200 | 0.206 |
| **Neural Network** | 0.561 | 0.800 | | 0.630 |
| **Logistic Regression** | 0.474 | 0.794 | 0.370 | |

**Table 5.** F1.

| | Tree | Naive Bayes | Neural Network | Logistic Regression |
|---|---|---|---|---|
| **Tree** | | 0.826 | 0.394 | 0.520 |
| **Naive Bayes** | 0.174 | | 0.205 | 0.222 |
| **Neural Network** | 0.606 | 0.795 | | 0.665 |
| **Logistic Regression** | 0.480 | 0.778 | 0.335 | |

**Table 6.** Precision.

| | Tree | Naive Bayes | Neural Network | Logistic Regression |
|---|---|---|---|---|
| **Tree** | | 0.347 | 0.307 | 0.475 |
| **Naive Bayes** | 0.434 | | 0.292 | 0.404 |
| **Neural Network** | 0.693 | 0.708 | | 0.741 |
| **Logistic Regression** | 0.525 | 0.596 | 0.259 | |

**Table 7.** Recall.

| | Tree | Naive Bayes | Neural Network | Logistic Regression |
|---|---|---|---|---|
| **Tree** | | 0.893 | 0.439 | 0.526 |
| **Naive Bayes** | 0.107 | | 0.200 | 0.206 |
| **Neural Network** | 0.561 | 0.800 | | 0.630 |
| **Logistic Regression** | 0.474 | 0.794 | 0.370 | |

**Table 8.** Specificity.

| | Tree | Naive Bayes | Neural Network | Logistic Regression |
|---|---|---|---|---|
| **Tree** | | 0.589 | 0.335 | 0.681 |
| **Naive Bayes** | 0.411 | | 0.215 | 0.475 |
| **Neural Network** | 0.665 | 0.785 | | 0.767 |
| **Logistic Regression** | 0.319 | 0.525 | 0.233 | |

**Table 9.** Logistic loss.

|  | Tree | Naive Bayes | Neural Network | Logistic Regression |
|---|---|---|---|---|
| **Tree** |  | 0.115 | 0.984 | 0.991 |
| **Naive Bayes** | 0.885 |  | 0.986 | 0.987 |
| **Neural Network** | 0.016 | 0.014 |  | 0.896 |
| **Logistic Regression** | 0.009 | 0.013 | 0.104 |  |

As illustrated in Table 3, the logistic regression model exhibits the highest AUC value (0.791), thereby indicating its superior overall capacity to discriminate between the various classes. The neural network exhibits a second-best AUC (0.740), yet this value is markedly inferior to that of the logistic regression model. The performance of the tree and naive Bayes models in terms of AUC indicates that they are less effective at distinguishing between the classes. Table 4 reveals that tree exhibits the highest classification accuracy (0.893), indicating that it is the most effective at correctly classifying instances. Moderate accuracies are observed for the neural network (0.561) and logistic regression (0.474). Conversely, naive Bayes has the lowest accuracy (0.107), suggesting poor performance in classifying data correctly. Table 5 reveals that the neural network has the highest F1-score (0.795), indicating an optimal balance between precision and recall. Logistic regression and tree also demonstrate satisfactory performance (0.665 and 0.520, respectively), whereas naive Bayes exhibits the lowest F1-score (0.174), suggesting suboptimal precision and recall balancing. Table 6 reveals that the neural network has the highest precision (0.741), indicating an ability to minimize false positives. Logistic regression also performs well with a precision of 0.525, while naive Bayes and tree demonstrate lower precision values, with naive Bayes exhibiting the lowest precision (0.292). Table 7 reveals that tree has the highest recall (0.893), indicating its effectiveness in identifying all relevant instances. The neural network and logistic regression also demonstrate high recall scores, although not as high as that of tree. Conversely, naive Bayes exhibits a notably low recall (0.200), suggesting a tendency to overlook relevant instances. Table 8 reveals that the neural network exhibits the highest specificity (0.767), indicating its efficacy in identifying negative instances. Both tree and logistic regression demonstrate satisfactory performance with specificity values of 0.681 and 0.319, respectively. Conversely, naive Bayes exhibits lower specificity values, suggesting its diminished capacity in identifying non-relevant instances. Finally, Table 9 reveals that the neural network and logistic regression models exhibit the lowest logistic loss values, indicating superior performance in terms of minimizing prediction error. In contrast, the tree and naive Bayes models demonstrate higher logistic loss, suggesting inferior performance in this regard.

Logistic regression excels in AUC and shows competitive results in F1-score, precision, and recall, making it a strong overall model. The neural network performs well in AUC, F1-score, precision, and specificity but struggles with classification accuracy and logistic loss. Tree shows high classification accuracy and recall but is weaker in AUC, precision, and logistic loss. Naive Bayes generally performs poorly across most metrics, including AUC, classification accuracy, precision, recall, and specificity.

Overall, logistic regression appears to be the most balanced model, offering strong performance across multiple metrics, while the neural network and tree have their own strengths and weaknesses. Naive Bayes is consistently the least effective across the board.

The confusion matrix gives the number or proportion of instances between the predicted and actual class. The selection of the elements in the matrix feeds the corresponding instances into the output signal. This way, one can observe which specific instances were misclassified and how. Figure 7 presents the confusion matrices of the neural network, tree, logistic regression, and naive Bayes models.

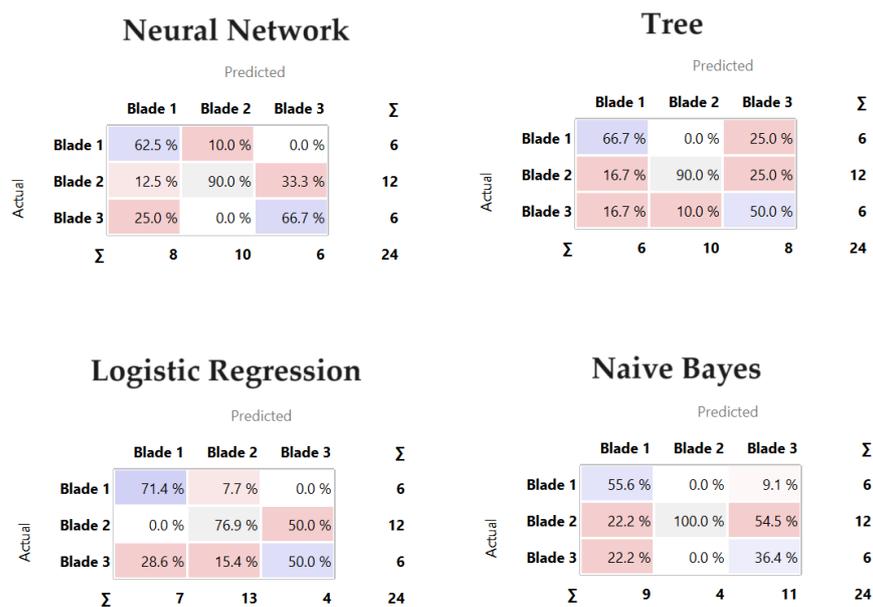

**Figure 7.** Confusion matrix.

The figure shows the confusion matrices for four different machine learning models: neural network, tree, logistic regression, and naive Bayes. In visual representations of confusion matrices, different colors are often used to make interpretation easier: Blue often used to highlight correct predictions (True Positive and True Negative); Red or Orange often used for incorrect predictions (False Positive and False Negative). Sometimes, the intensity of the color indicates the magnitude of the values, with darker shades representing higher numbers. Each confusion matrix illustrates the performance of these models in predicting categories labeled as "Blade 1", "Blade 2", and "Blade 3".

Confusion Matrices Overview: Rows (Actual): Represent the true classes; Columns (Predicted): Represent the classes predicted by the model; and Cells: Show the percentage of correct or incorrect predictions for each class.

Models Analysis

Neural Network:

Blade 1: 62.5% correctly classified, 10% misclassified as Blade 2, 0% as Blade 3.

Blade 2: 90% correctly classified, some misclassification to Blade 1 and Blade 3.

Blade 3: 66.7% correctly classified, 25% misclassified as Blade 1.

The Neural Network shows strong performance in predicting Blade 2 and Blade 3 but slightly lower accuracy for Blade 1.

Tree:

Blade 1: 66.7% correctly classified.

Blade 2: 90% correctly classified.

Blade 3: 50% correctly classified.

The decision tree model performs well with Blade 1 and Blade 2 but has some difficulty with Blade 3.

Logistic Regression:

Blade 1: 71.4% correctly classified.

Blade 2: 76.9% correctly classified.

Blade 3: 50% correctly classified.

Logistic regression shows the best accuracy for Blade 1 but still struggles with Blade 3.

Naive Bayes:

Blade 1: 55.6% correctly classified.

Blade 2: 100% correctly classified.

Blade 3: 54.5% correctly classified.

Naive Bayes has perfect accuracy for Blade 2 but lower performance for Blade 1 and Blade 3.

Blade 2: All models perform quite well, with naive Bayes achieving 100% accuracy.

Blade 1 and Blade 3: Generally, show lower performance across all models, with the neural network and logistic regression performing slightly better.

Naive Bayes: Best at predicting Blade 2 but less accurate with Blade 1 and Blade 3.

The results are shown in Figure 8, where the algorithm's selection is displayed, grouping images that highlight the damage to the wind turbine blades.

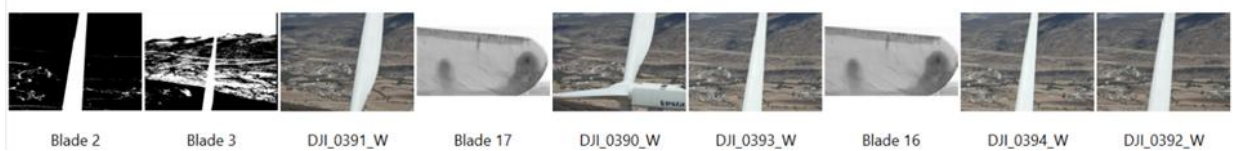

**Figure 8.** Wind turbine blades selected.

### 3.2. Image Clustering

Once the image embedding was performed, the Distances tool was used, and the resulting distance matrix could be fed further into hierarchical clustering to uncover groups in the data.

The result of the hierarchical clustering is presented in Figure 9.

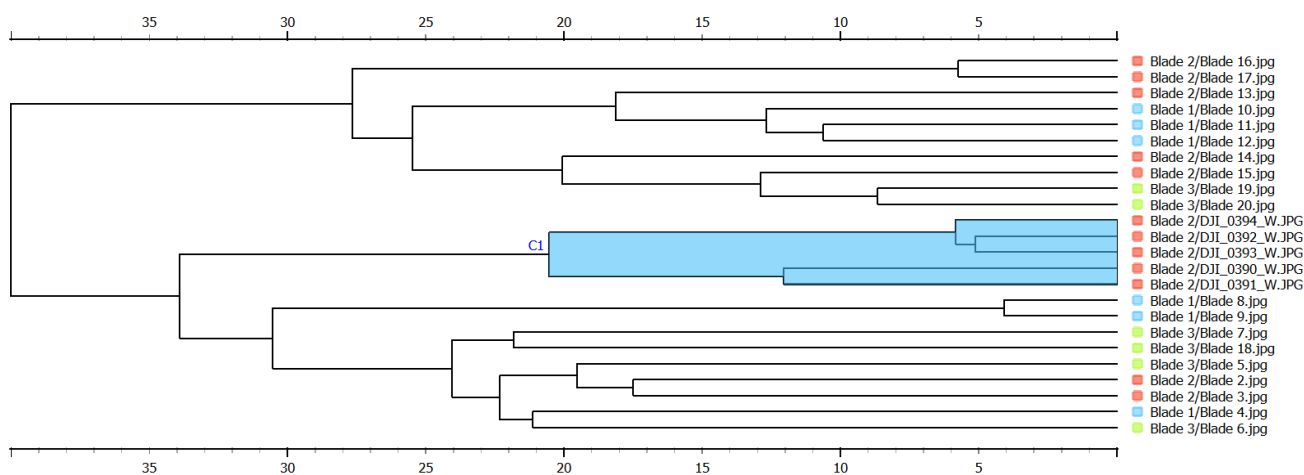

**Figure 9.** Hierarchical clustering of wind blade images.

Figure 9 shows the results of a hierarchical clustering analysis applied to a set of images of wind turbine blades with defects. The horizontal axis represents the distance or dissimilarity between clusters as they merge. A higher value indicates a greater difference between clusters before they are combined. The vertical axis displays the names of the blade images. The images are grouped into different branches, where each branch fusion represents the combination of two clusters into one. Mergers occur from individual elements (the images) to the final cluster that contains all the data.

By observing Figure 9, there are some well-defined groups, such as the images "Blade 3/Blade 5.jpg" and "Blade 3/Blade 4.jpg," which are clustered very closely together, indicating they are like each other based on the characteristics considered in the analysis.

The image labels are colored, which may indicate different predefined categories or subgroups of blades (perhaps based on defect type or blade location).

By examining the structure of the branches, the groups of images with similar colors tend to cluster together, suggesting that the images within the same color group share similar characteristics.

Figure 10 shows the Image Viewer widget, which was used to display the results of hierarchical clustering.

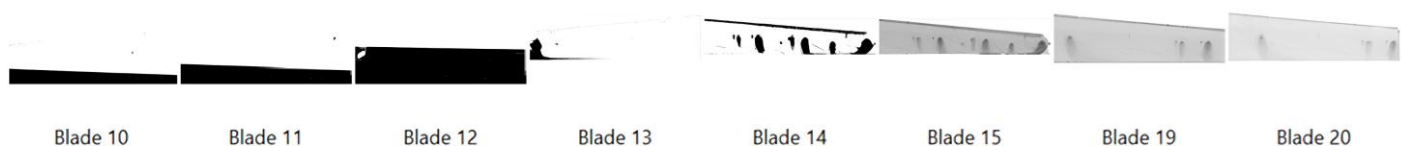

**Figure 10.** Image Viewer used to display results of hierarchical clustering.

By carrying out both procedures for recognizing damage in wind turbine blades, it is observed that the hierarchical clustering model performed better than the logistic regression model, as it grouped the images more effectively with fewer computational resources.

## 4. Discussion

This study presents advancements in fault detection methodologies for wind turbine blades, a critical component in the wind energy sector [50]. Given the increasing reliance on wind power as a sustainable energy source, the operational efficiency and lifespan of wind turbines have become paramount. Wind turbine blades, exposed to adverse environmental conditions and dynamic mechanical loads, are particularly susceptible to a range of faults that can significantly impact energy production efficiency. Therefore, early detection of such faults is essential for maintaining system reliability, reducing downtime, and minimizing maintenance costs.

The study's dual-model approach, employing both logistic regression and clustering techniques, offers a novel perspective on fault detection, contrasting with the predominantly supervised methods reported in existing literature. The performance of logistic regression, surpassing that of neural networks, decision trees, and naive Bayes in this context, challenges the prevailing assumption that more complex models inherently provide better results in fault detection scenarios. This finding aligns with recent trends in machine learning, where simpler models are sometimes preferred for their interpretability and efficiency, particularly when data characteristics do not support more intricate architectures.

The superior performance of the clustering model in data segmentation highlights the potential of unsupervised learning techniques to capture the underlying patterns of faults in the blades. This result is particularly significant as it suggests that traditional supervised methods may overlook certain fault patterns that become more apparent when data are analyzed without predefined labels. The study contributes to the growing body of work exploring the effectiveness of clustering in industrial applications, particularly in cases where labeled data are scarce or costly to obtain.

The use of Orange Data Mining as a practical tool for implementing these models underscores the accessibility of advanced computational learning techniques. The ability to leverage such user-friendly platforms without sacrificing analytical rigor opens the door for broader adoption of these methodologies in the industry. This democratization of data science tools is crucial for smaller organizations or those with limited resources, which can benefit from sophisticated fault detection systems without the need for extensive computational infrastructure or specialized expertise.

The findings suggest that clustering-based models, due to their capacity to discern subtle fault patterns, may be more effective in environments where data heterogeneity is high. However, the study also raises important questions about the balance between model complexity and interpretability, especially in critical applications such as wind turbine maintenance where the stakes are high.

While the study successfully demonstrates the utility of both logistic regression and clustering in fault detection, several challenges remain. The performance of these models in real-world scenarios, where data quality and availability can vary significantly, warrants further investigation. Additionally, integrating these models into existing supervisory control and data acquisition (SCADA) systems presents both technical and logistical challenges that must be addressed for practical implementation.

Future research could focus on hybrid approaches that combine the strengths of both supervised and unsupervised learning. For instance, semi-supervised learning techniques could be explored to leverage available labeled data while also capturing the unstructured patterns detected by clustering. Furthermore, the study's methodology could be extended to other components of wind turbines, such as gearboxes or generators, where similar fault detection challenges exist.

## 5. Conclusions

The methodology proposed in this study for detecting faults in wind turbine blades using computational learning techniques has proven to be both effective and robust. The use of logistic regression as part of the first model outperformed neural networks, decision trees, and the naive Bayes method, highlighting its value as a tool for identifying patterns associated with faults. However, the clustering approach implemented in the second model demonstrated notable superiority in both precision and data segmentation capability, suggesting that clustering can better capture the underlying characteristics of the data compared to supervised methods.

These findings underscore the importance of considering multiple approaches within computational learning for fault detection in complex systems such as wind turbine blades. The methodology presented not only offers a new perspective on early fault identification but also emphasizes the utility of accessible and flexible tools like Orange Data Mining for implementing advanced solutions in the wind energy industry. Future research could focus on the combination of these methods to further enhance the accuracy and efficiency of the detection system, as well as on applying these techniques to other critical components within the energy infrastructure.

Further work is required to combine methods and apply these techniques to other critical components in energy infrastructure, such as hybrid modeling approaches and the combination of different models, including neural networks, decision trees, and logistic regression, into an ensemble approach such as stacking, bagging, or boosting. This could enhance predictive performance. Additionally, the implementation of a voting mechanism, whereby the predictions of multiple models are combined to arrive at a final decision, has the potential to enhance the accuracy of the process. The models could be assigned weights based on their performance on specific classes, such as Blade 2 in this study. Furthermore, multiple neural networks with varied architectures or training data subsets could be combined. Techniques such as dropout could also be employed to create an ensemble effect within a single neural network model. The application of pre-trained models from similar domains and fine-tuning of this specific problem could facilitate model development and enhance accuracy, particularly in cases with limited data. Additionally, the exploration of advanced feature engineering techniques to create more discriminative features could assist in improving the model's performance. Techniques such as principal component analysis (PCA) for dimensionality reduction or the utilization of domain knowledge to engineer novel features pertinent to blade failures may prove beneficial. The development of techniques to detect outliers or anomalies that may indicate the early signs of failure could serve to enhance the efficacy of the classification models. Anomalies may be identified through the utilization of clustering techniques, statistical methods, or specialized models such as autoencoders. Furthermore, the techniques explored herein could be adapted for the prediction of failures in other critical components, including wind turbine blades and solar panel systems. These components exhibit analogous characteristics with respect to wear and tear, environmental impact, and importance in energy generation.

**Author Contributions:** Conceptualization, J.A.-A., D.C.G.C., L.A.S.-L., D.R.-R., J.A.F., A.-J.P.-M., D.M.-R. and Q.H.-E.; methodology, J.A.-A., D.C.G.C., D.R.-R., J.A.F., A.-J.P.-M., D.M.-R. and Q.H.-E.; software, L.A.S.-L., D.R.-R., J.A.F., A.-J.P.-M. and Q.H.-E.; formal analysis, L.A.S.-L., D.R.-R., J.A.F., A.-J.P.-M. and Q.H.-E.; investigation, L.A.S.-L., D.R.-R., J.A.F., A.-J.P.-M., D.M.-R. and Q.H.-E.; writing—original draft preparation, L.A.S.-L., D.R.-R., J.A.F., A.-J.P.-M. and Q.H.-E.; writing—review and editing, L.A.S.-L., D.R.-R., J.A.F., A.-J.P.-M., D.M.-R. and Q.H.-E.; visualization, L.A.S.L., D.R.-R., J.A.F., A.-J.P.-M. and Q.H.-E.; supervision, L.A.S.-L., D.R.-R., J.A.F., A.-J.P.-M. and Q.H.-E. All authors have read and agreed to the published version of the manuscript.

**Funding:** This research received no external funding.

**Institutional Review Board Statement:** Not applicable.

**Informed Consent Statement:** Not applicable

**Data Availability Statement:** The original contributions presented in the study are included in the article, further inquiries can be directed to the corresponding authors.

**Funding:** This research was funded by Universidad Nacional Autónoma de México through the project "Determinación de fallas en álabes de aerogeneradores a través de procesamiento digital de imágenes y aprendizaje automático no supervisado", grant number IA 100023.

**Acknowledgments:** The authors express their acknowledgment to the Universidad Nacional Autónoma de México through the project PAPIIT TA100323 "Diseño e implementación de un banco de pruebas experimental para el estudio de sistemas de control de flujo activo en aerogeneradores a escala; una perspectiva hacía el desarrollo de aspas inteligentes". We are grateful for the academic enthusiasm of the students of the Renewable Energy Engineering program at ENES Juriquilla—UNAM.

**Conflicts of Interest:** The authors declare no conflicts of interest.

## References

1. Tweneboah-Koduah, D.; Arah, M.; Botchway, T. Globalization, Renewable Energy Consumption and Sustainable Development. *Cogent Soc. Sci.* **2023**, *9*, 2223399. https://doi.org/10.1080/23311886.2023.2223399.
2. Wang, B.; Li, Y.; Gao, S.; Shen, K.; Zhao, S.; Yao, Y.; Zhou, Z.; Hu, Z.; Zheng, X. Stability Analysis of Wind Turbine Blades Based on Different Structural Models. *J. Mar. Sci. Eng.* **2023**, *11*, 1106. https://doi.org/10.3390/jmse11061106.
3. Deng, L.; Liu, S.; Shi, W.; Xu, J. Defect Detection and Classification of Offshore Wind Turbine Rotor Blades. *Nondestruct. Test. Eval.* **2023**, *39*, 954–975. https://doi.org/10.1080/10589759.2023.2234554.
4. Somaiday, B.; Czajka, I.; Yass, M. Strenght Analysis of a Blade with Different Cross-Section. *Facta Univ.-Ser. Electron. Energetics* **2023**, *36*, 239–251. https://doi.org/10.2298/FUEE2302239S.
5. Ciang, C.C.; Lee, J.-R.; Bang, H.-J. Structural Health Monitoring for a Wind Turbine System: A Review of Damage Detection Methods. *Meas. Sci. Technol.* **2008**, *19*, 122001.
6. Zhao, Q.; Yuan, Y.; Sun, W.; Fan, X.; Fan, P.; Ma, Z. Reliability Analysis of Wind Turbine Blades Based on Non-Gaussian Wind Load Impact Competition Failure Model. *Measurement* **2020**, *164*, 107950. https://doi.org/10.1016/j.measurement.2020.107950.
7. Amirat, Y.; Choqueuse, V.; Benbouzid, M. EEMD-Based Wind Turbine Bearing Failure Detection Using the Generator Stator Current Homopolar Component. *Mech. Syst. Signal Process.* **2013**, *41*, 667–678. https://doi.org/10.1016/j.ymssp.2013.06.012.
8. de Novaes Pires Leite, G.; Farias, F.C.; de Sá, T.G.; Araújo da Costa, A.C.; Petribú Brennand, L.J.; de Souza, M.G.G.; Villa, A.A.O.; Droguett, E.L. A Robust Fleet-Based Anomaly Detection Framework Applied to Wind Turbine Vibration Data. *Eng. Appl. Artif. Intell.* **2023**, *126*, 106859. https://doi.org/10.1016/j.engappai.2023.106859.
9. Rizk, P.; Younes, R.; Ilinca, A.; Khoder, J. Wind Turbine Blade Defect Detection Using Hyperspectral Imaging. *Remote Sens. Appl. Soc. Environ.* **2021**, *22*, 100522. https://doi.org/10.1016/j.rsase.2021.100522.
10. Han, H.; Yang, D. Correlation Analysis Based Relevant Variable Selection for Wind Turbine Condition Monitoring and Fault Diagnosis. *Sustain. Energy Technol. Assess.* **2023**, *60*, 103439. https://doi.org/10.1016/j.seta.2023.103439.
11. Zhang, Y.; Qin, X.; Han, Y.; Huang, Q. Vibration Amplitude Normalization Enhanced Fault Diagnosis under Conditions of Variable Speed and Extremely Limited Samples. *Meas. Sci. Technol.* **2023**, *34*, 125111. https://doi.org/10.1088/1361-6501/aced4e.
12. Han, T.; Xie, W.; Pei, Z. Semi-Supervised Adversarial Discriminative Learning Approach for Intelligent Fault Diagnosis of Wind Turbine. *Inf. Sci.* **2023**, *648*, 119496. https://doi.org/10.1016/j.ins.2023.119496.
13. Wang, L.; He, Y.; Shao, K.; Xing, Z.; Zhou, Y. An Unsupervised Approach to Wind Turbine Blade Icing Detection Based on Beta Variational Graph Attention Autoencoder. *IEEE Trans. Instrum. Meas.* **2023**, *73*, 1–12. https://doi.org/10.1109/TIM.2023.3286011.

14. Guo, J.; Liu, C.; Cao, J.; Jiang, D. Damage Identification of Wind Turbine Blades with Deep Convolutional Neural Networks. *Renew. Energy* **2021**, *174*, 122–133. https://doi.org/10.1016/j.renene.2021.04.040.

15. Bernalte Sánchez, P.J.; García Márquez, F.P. 4—Artificial Neural Networks Applied for Wind Turbines Maintenance Management in Unmanned Aerial Vehicle Acoustic Inspection Case. In *Non-Destructive Testing and Condition Monitoring Techniques in Wind Energy*; Garcia Marquez, F.P., Papaelias, M., Junior, V.L.J., Eds.; Academic Press: Cambridge, MA, USA, 2023; pp. 37–49, ISBN 978-0-323-99666-2.

16. Liu, Z.; Zhang, L. Naturally Damaged Wind Turbine Blade Bearing Fault Detection Using Novel Iterative Nonlinear Filter and Morphological Analysis. *IEEE Trans. Ind. Electron.* **2020**, *67*, 8713–8722. https://doi.org/10.1109/TIE.2019.2949522.

17. Li, W.; Pan, Z.; Hong, N.; Du, Y. Defect Detection of Large Wind Turbine Blades Based on Image Stitching and Improved Unet Network. *J. Renew. Sustain. Energy* **2023**, *15*, 013302. https://doi.org/10.1063/5.0125563.

18. Wang, Z.; Xu, J.; Jia, Y.; Cai, C.; Zhou, T.; Wang, X.; Xu, J.; Li, Q. A New Fault Detection Strategy for Wind Turbine Rotor Imbalance Based on Multi-Condition Vibration Signal Analysis. *J. Renew. Sustain. Energy* **2023**, *15*, 033307. https://doi.org/10.1063/5.0149852.

19. Banala, H.S.; Sahoo, S.; Sethi, M.R.; Sharma, A.K. Fault Diagnosis in Wind Turbine Blades Using Machine Learning Techniques. In *Machine Learning, Image Processing, Network Security and Data Sciences*; Springer: Singapore, 2023; Volume 946, pp. 401–411.

20. Ogaili, A.A.F.; Hamzah, M.N.; Jaber, A.A. Integration of Machine Learning (ML) and Finite Element Analysis (FEA) for Predicting the Failure Modes of a Small Horizontal Composite Blade. *Int. J. Renew. Energy Res.* **2022**, *12*, 2168–2179. https://doi.org/10.20508/ijrer.v12i4.13354.g8589.

21. Rangel-Rodriguez, A.H.; Huerta-Rosales, J.R.; Amezquita-Sanchez, J.P.; Granados-Lieberman, D.; Bueno-Lopez, M.; Valtierra-Rodriguez, M. Detection of Multiple Faults in a Low-Power Wind Turbine by Using Convolutional Neural Networks. In Proceedings of the 2022 IEEE International Autumn Meeting on Power, Electronics and Computing (ROPEC), Ixtapa, Mexico, 9–11 November 2022.

22. Gajbhiye, A.; Warudkar, V. Convolution Neural Network for Structural Failure Detection of Wind Turbine Blade: A Review. In *Advances in Mechanical Engineering and Technology*; Springer: Singapore, 2022; pp. 467–473.

23. Wang, Y.; Zhu, C.; Li, Y.; Tan, J. The Effect of Reduced Power Operation of Faulty Wind Turbines on the Total Power Generation for Different Wind Speeds. *Sustain. Energy Technol. Assess.* **2021**, *45*, 101178. https://doi.org/10.1016/j.seta.2021.101178.

24. Zhang, J.; Kang, J.; Sun, L.; Bai, X. Risk Assessment of Floating Offshore Wind Turbines Based on Fuzzy Fault Tree Analysis. *Ocean. Eng.* **2021**, *239*, 109859. https://doi.org/10.1016/j.oceaneng.2021.109859.

25. Mourad, A.-H.I.; Almomani, A.; Ahmad Sheikh, I.; Elsheikh, A.H. Failure Analysis of Gas and Wind Turbine Blades: A Review. *Eng. Fail. Anal.* **2023**, *146*, 107107. https://doi.org/10.1016/j.engfailanal.2023.107107.

26. Du, B.; Narusue, Y.; Furusawa, Y.; Nishihara, N.; Indo, K.; Morikawa, H.; Iida, M. Clustering wind turbines for SCADA data-based fault detection. *IEEE Trans. Sustain. Energy* **2022**, *14*, 442–452. https://doi.org/10.1109/TSTE.2022.3215672 .

27. Choe Wei Chang, C.; Jian Ding, T.; Han, W.; Chong Chai, C.; Bhuiyan, M.A.S.; Choon-Yian, H.; Chuan Song, T. Recent Advancements in Condition Monitoring Systems for Wind Turbines: A Review. *Energy Rep.* **2023**, *9*, 22–27. https://doi.org/10.1016/j.egyr.2023.08.061.

28. Ogaili, A.A.F.; Abdulhady Jaber, A.; Hamzah, M.N. Wind Turbine Blades Fault Diagnosis Based on Vibration Dataset Analysis. *Data Brief* **2023**, *49*, 109414. https://doi.org/10.1016/j.dib.2023.109414.

29. Kong, K.; Dyer, K.; Payne, C.; Hamerton, I.; Weaver, P.M. Progress and Trends in Damage Detection Methods, Maintenance, and Data-Driven Monitoring of Wind Turbine Blades—A Review. *Renew. Energy Focus* **2023**, *44*, 390–412. https://doi.org/10.1016/j.ref.2022.08.005.

30. Yang, X.; Ma, D.; Zhang, L.; Yu, Y.; Yao, Y.; Yang, M. High-Fidelity Multi-Level Efficiency Optimization of Propeller for High Altitude Long Endurance UAV. *Aerosp. Sci. Technol.* **2023**, *133*, 108142.

31. Wang, F.; Zhang, C.; Zhang, W.; Fang, C.; Xia, Y.; Liu, Y.; Dong, H. Object-Based Reliable Visual Navigation for Mobile Robot. *Sensors* **2022**, *22*, 2387.

32. Wang, Y. 3D Dynamic Image Modeling Based on Machine Learning in Film and Television Animation. *J. Multimed. Inf. Syst.* **2023**, *10*, 69–78.

33. Lei, Y.; Yang, B.; Jiang, X.; Jia, F.; Li, N.; Nandi, A.K. Applications of Machine Learning to Machine Fault Diagnosis: A Review and Roadmap. *Mech. Syst. Signal Process.* **2020**, *138*, 106587.

34. El-Hajj, C.; Kyriacou, P.A. A Review of Machine Learning Techniques in Photoplethysmography for the Non-Invasive Cuff-Less Measurement of Blood Pressure. *Biomed. Signal Process. Control* **2020**, *58*, 101870.

35. Ciuriuc, A.; Rapha, J.I.; Guanche, R.; Domínguez-García, J.L. Digital Tools for Floating Offshore Wind Turbines (FOWT): A State of the Art. *Energy Rep.* **2022**, *8*, 1207–1228. https://doi.org/10.1016/j.egyr.2021.12.034.

36. Barbot, A.; Gatti, R. Unsupervised Learning for Structure Detection in Plastically Deformed Crystals. *Comput. Mater. Sci.* **2023**, *230*, 112459. https://doi.org/10.1016/j.commatsci.2023.112459.

37. Blumenthal, J.; Megherbi, D.B.; Lussier, R. Unsupervised Machine Learning via Hidden Markov Models for Accurate Clustering of Plant Stress Levels Based on Imaged Chlorophyll Fluorescence Profiles & Their Rate of Change in Time. *Comput. Electron. Agric.* **2020**, *174*, 105064.

38. Matteucci, G.; Piasini, E.; Zoccolan, D. Unsupervised Learning of Mid-Level Visual Representations. *Curr. Opin. Neurobiol.* **2024**, *84*, 102834. https://doi.org/10.1016/j.conb.2023.102834.

39. Wang, Y.; Lu, Q.; Ren, B. Wind Turbine Crack Inspection Using a Quadrotor with Image Motion Blur Avoided. *IEEE Robot. Autom. Lett.* **2023**, *8*, 1069–1076. https://doi.org/10.1109/LRA.2023.3236576.
40. van Leeuwen, R.; Koole, G. Data-Driven Market Segmentation in Hospitality Using Unsupervised Machine Learning. *Mach. Learn. Appl.* **2022**, *10*, 100414. https://doi.org/10.1016/j.mlwa.2022.100414.
41. Hurtik, P.; Molek, V.; Perfilieva, I. Novel Dimensionality Reduction Approach for Unsupervised Learning on Small Datasets. *Pattern Recognit.* **2020**, *103*, 107291. https://doi.org/10.1016/j.patcog.2020.107291.
42. Thenier-Villa, J.L.; Martínez-Ricarte, F.R.; Figueroa-Vezirian, M.; Arikan-Abelló, F. Glioblastoma Pseudoprogression Discrimination Using Multiparametric Magnetic Resonance Imaging, Principal Component Analysis, Supervised and Unsupervised Machine Learning. *World Neurosurg.* **2024**, *183*, e953–e962. https://doi.org/10.1016/j.wneu.2024.01.074.
43. Mohapatra, S.; Swarnkar, T. *Comparative Study of Different Orange Data Mining Tool-Based AI Techniques in Image Classification*; Springer: Berlin/Heidelberg, Germany, 2021; pp. 611–620.
44. Ishak, A.; Siregar, K.; Ginting, R.; Afif, M. *Orange Software Usage in Data Mining Classification Method on the Dataset Lenses*; IOP Publishing: Bristol, UK, 2020; Volume 1003, p. 012113.
45. Tebala, D.; Marino, D. Companies and Artificial Intelligence: An Example of Clustering with Orange. In *Innovations and Economic and Social Changes due to Artificial Intelligence: The State of the Art*; Springer: Berlin/Heidelberg, Germany, 2023; pp. 1–12.
46. Karuppasamy, A.; Abdesselam, A.; Hedjam, R.; zidoum, H.; Al-Bahri, M. Feed-Forward Networks Using Logistic Regression and Support Vector Machine for Whole-Slide Breast Cancer Histopathology Image Classification. *Intell.-Based Med.* **2024**, *9*, 100126. https://doi.org/10.1016/j.ibmed.2023.100126.
47. Li, Z.; Yang, W.; Qi, H.; Jin, L.; Huang, Y.; Ding, K. A Tree-Based Model with Branch Parallel Decoding for Handwritten Mathematical Expression Recognition. *Pattern Recognit.* **2024**, *149*, 110220. https://doi.org/10.1016/j.patcog.2023.110220.
48. Hubert; Phoenix, P.; Sudaryono, R.; Suhartono, D. Classifying Promotion Images Using Optical Character Recognition and Naïve Bayes Classifier. *Procedia Comput. Sci.* **2021**, *179*, 498–506. https://doi.org/10.1016/j.procs.2021.01.033.
49. Liu, Y.; Xue, J.; Li, D.; Zhang, W.; Chiew, T.K.; Xu, Z. Image Recognition Based on Lightweight Convolutional Neural Network: Recent Advances. *Image Vis. Comput.* **2024**, *146*, 105037. https://doi.org/10.1016/j.imavis.2024.105037.
50. Li, W.; Zhang, L.; Wu, C.; Cui, Z.; Niu, C. A New Lightweight Deep Neural Network for Surface Scratch Detection. *Int. J. Adv. Manuf. Technol.* **2022**, *123*, 1999–2015. https://doi.org/10.1007/s00170-022-10335-8.